\documentclass{article}

\usepackage{PRIMEarxiv}

\usepackage[utf8]{inputenc} 
\usepackage[T1]{fontenc}    
\usepackage{hyperref}       
\usepackage{url}            
\usepackage{booktabs}       
\usepackage{amsfonts}       
\usepackage{nicefrac}       
\usepackage{microtype}      
\usepackage{lipsum}
\usepackage{fancyhdr}       
\usepackage{graphicx}       
\graphicspath{{media/}}     

\usepackage{amsmath}

\usepackage{algorithm}
\usepackage{algorithmic}
\usepackage{multirow}

\title{Smart Routing for EV Charge Point Operators in Mega Cities: Case Study of Istanbul}

\author{
  Onur Yenigun,  \\
  Department of Computer Engineering \\
  Marmara University \\
  Istanbul, Turkiye\\
  \texttt{onuryenigun@marun.edu.tr} \\
   \And
  Gozde Karatas Baydogmus \\
  Department of Computer Science \\
  Loyola University of Chicago \\
  Chicago, USA\\
  \texttt{gkaratasbaydogmus@luc.edu} \\
  \And
  Kazim Yildiz \\
  Department of Computer Engineering \\
  Marmara University \\
  Istanbul, Turkiye\\
  \texttt{kazim.yildiz@marmara.edu.tr} \\ \\
}

\begin{document}
\maketitle

\begin{abstract}
The rapidly increasing use of electric vehicles (EVs) has made it even more important to manage the charging infrastructure sustainably. The expansion of charging station networks, especially in large cities, creates serious logistical challenges for charging point operators (CPOs) in planning maintenance and repair activities. Inefficient field personnel management can lead to time loss, high operational costs, and resource waste. This study presents an integrated method to optimize the planning of EV charging network maintenance operations. The proposed approach groups charging stations according to geographical proximity using the K-means clustering algorithm and calculates the shortest routes between clusters using a genetic algorithm. The method was developed in Python and applied to a dataset consisting of 100 EV charging stations in Istanbul.

Considering the population density, traffic density, and resource constraints of Istanbul, the route planning approach presented in this study has great potential, especially for such metropolises. According to the different parameter configurations tested, the most efficient scenario provided approximately 35\% distance savings compared to the reference route created according to the sequential data layout. While the reference route provides a simple comparison, the study presents a solution that will enable field operations in metropolitan cities such as Istanbul to be conducted in a more efficient, planned and scalable manner. In future studies, it is planned to integrate real-time factors such as traffic conditions and field technician constraints.
\end{abstract}


\section{Introduction}
The proliferation of electric vehicles not only offers an environmentally friendly transportation option in giant metropolises like Istanbul; it also creates new and complex problems in terms of the management of urban infrastructures. The most important of these problems is the effective monitoring, maintenance and sustainable operation of rapidly growing charging station networks. Although the increasing number of charging stations facilitates user access, it also makes processes such as maintenance, repair and operational management more costly and complex.

\begin{figure}[htp]
        	\centering
        	\scalebox{0.40}{\includegraphics{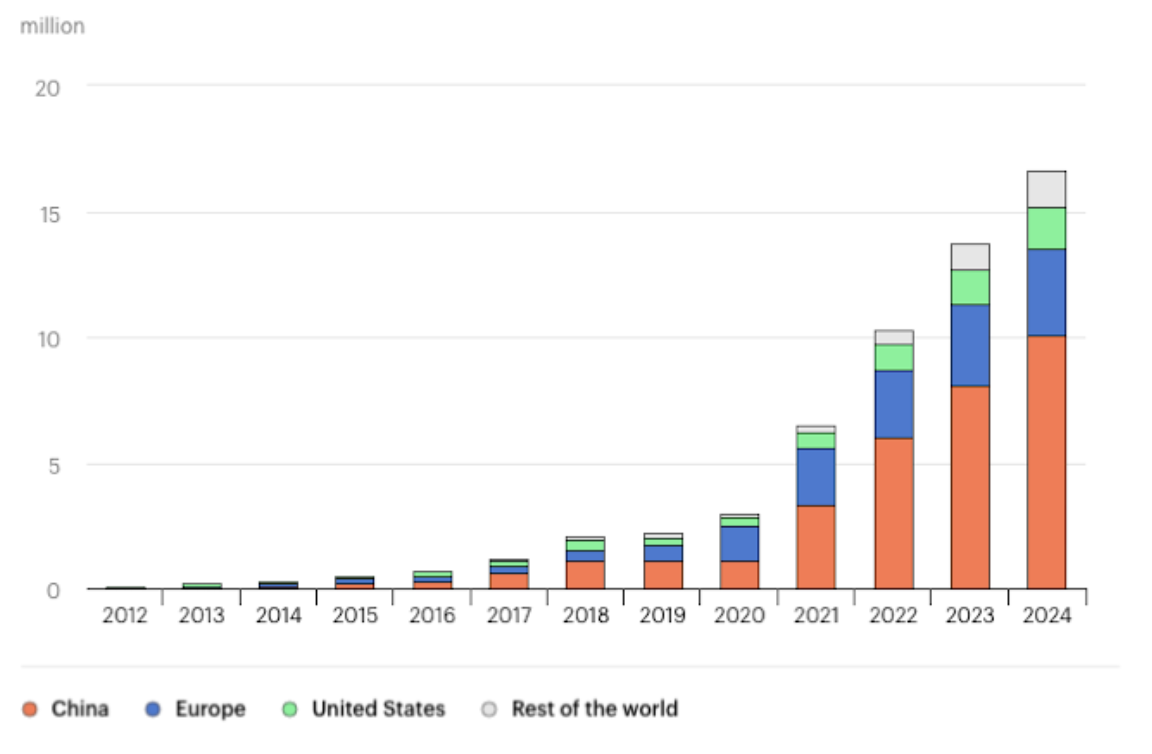}}
        	\caption{Global electric car sales data}
            \label{fig:figure1}
\end{figure}

The accelerating electric vehicle transition on a global scale has increased the importance of sustainable and efficient operational strategies in the fields of energy and e-mobility. In this context, charge point operators (CPO) must conduct their field operations in a more planned, time-efficient and resource-sensitive manner in order to maintain high service quality \cite{Vasiliki2023, Akbari2018}. According to data from the International Energy Agency (IEA) for 2024 which is given in Figure~\ref{fig:figure1} , the number of electric vehicles is set to account for approximately 18\% of all cars sold in 2023, compared to 14\% in 2022 and just 2\% in 2018, compared to 5 years ago. These trends suggest that growth remains strong as electric car markets mature \cite{iea_ev_2025}.

This study proposes a method that optimizes route planning for maintenance and repair services, using the example of Istanbul. The K-means algorithm was used to cluster charging stations according to geographical proximity, and the genetic algorithm was used to determine the shortest routes between clusters. With this combination, it was aimed to reduce the total travel distance of maintenance teams and achieve significant gains in terms of both time and cost \cite{Masoud2018}.

The presented approach is important in terms of showing how theoretical algorithms can be adapted to complex city conditions such as Istanbul. The results obtained show that this method is applicable not only mathematically but also practically; it contributes to CPOs making more sustainable operational decisions with map-based guidance \cite{Kumar2024}. These optimization approaches provide both mathematical and practical solutions to time and cost issues crucial for operators \cite{Kumar2024}. Maps and route recommendations generated as outputs of the study serve as a comprehensive guide for field operations, contributing to the more sustainable and cost-effective management of business processes. By enabling operators to access charging stations more quickly and systematically, this method plays a critical role in ensuring uninterrupted access to a secure and accessible charging network for electric vehicle users. Considering these aspects, the study aimed to address the following questions:

\begin{itemize}
    \item How can maintenance and repair operations of electric vehicle charging stations be made more efficient for large and dense cities?
    \item How can known methods such as K-means clustering and genetic algorithm be integrated into real field conditions in complex cities such as Istanbul?
    \item What level of improvement does route planning obtained with algorithmic optimization provide in terms of operational distance and resource usage?
\end{itemize}

The study, structured around these questions, presents a model that has not only theoretical but also practical application potential and has been tested for applicability at the metropolitan level. In the study, literature review is given in section 2; in section 3, the methodology used in the development of the study and the proposed model are explained; in section 4, experimental results are given; in section 5, the experimental results are discussed, and in section 6, conclusion and future work is given.

\section{Literature Review}
In recent years, numerous studies have addressed the optimization of electric vehicle (EV) infrastructure through various heuristic and metaheuristic techniques. These works have primarily focused on problems such as station location planning, routing optimization, and clustering to improve service quality, reduce costs, and enhance system efficiency.

Lazari \cite{Vasiliki2023} developed a multi-objective optimization model using a genetic algorithm to determine the optimal number and locations of charging stations. The model aims to balance cost minimization with service quality, offering valuable insights into strategic deployment under real-world constraints. Similarly, Karakatic \cite{optimize_non_linear_ev_routing} proposed a Two-Layer Genetic Algorithm to solve the Multi-Depot Vehicle Routing Problem with Time Windows (MDVRPTW) specifically tailored for EVs. This approach considers nonlinear charging durations and demonstrates effective performance in complex routing scenarios involving electric vehicles.

Efthymiou et al. \cite{ev_charging_infra_location} addressed the EV Charging Infrastructure Location Problem using genetic algorithms, relying on origin-destination data to forecast EV penetration. Their method supports optimized charger placement in urban areas, providing a data-driven framework for future infrastructure planning. Zhou et al. \cite{location_opt_ev_stations} further contributed to the location optimization literature by incorporating construction, operational, and environmental cost models into a GA-based algorithm. Their work highlights cost-sensitive network expansion strategies, particularly relevant for public infrastructure projects.

In the area of clustering and algorithmic integration, Irfan et al. \cite{opt_kmeans_ga} demonstrated that combining K-means with genetic algorithms can improve convergence speed and clustering stability. Zeebaree et al. \cite{kmeans_ga_review} conducted a broader review of this hybridization, emphasizing its value in minimizing distance and computational time—critical in routing and distribution problems. Other researchers have also explored advanced hybrid clustering methods. Komarasamy et al. \cite{optimized_kmeans_bat} used the Bat Algorithm to address K-means’ limitations, while Roy et al. \cite{genetic_kmeans_mixed} extended genetic K-means methods to handle mixed-type datasets, improving adaptability across domains. Table~\ref{tab:table1} provides a comparison of existing studies in the literature.

\renewcommand{\arraystretch}{1.4} 
\begin{table}[htbp]
\centering
\caption{Comparison of the literature papers}
\label{tab:table1}
\resizebox{\linewidth}{!}{
\begin{tabular}{@{}llllll@{}}
\toprule
\textbf{Study} & \textbf{Method(s) Used} & \textbf{Focus Area} & \textbf{Real Data} & \textbf{Real City Application} & \textbf{Key Difference / Contribution} \\ \midrule
\cite{Vasiliki2023} & Multi-objective Genetic Algorithm & Charging station location planning & Yes & No & Focuses on location siting, not routing \\
\cite{optimize_non_linear_ev_routing} & Two-layer Genetic Algorithm & Routing with nonlinear charging times & Partially & No & Time window, recharge constraints \\
\cite{ev_charging_infra_location} & Genetic Algorithm & Location planning with O-D data & Yes & No & Based on transport behavior data \\
\cite{location_opt_ev_stations} & Genetic Algorithm + Cost Model & Construction and operational cost minimization & Yes & No & Cost-oriented location optimization \\
\cite{opt_kmeans_ga} & Genetic Algorithm + K-means & Improved clustering convergence & No & No & Focus on algorithm efficiency, not application \\
\cite{kmeans_ga_review} & Genetic Algorithm + K-means & Distance reduction in clustering & No & No & General system evaluation, no real-world context \\
\cite{optimized_kmeans_bat} & K-means + Bat Algorithm & Enhanced clustering performance & No & No & Tackles sensitivity to initial centroids \\
\cite{genetic_kmeans_mixed} & Genetic K-means & Mixed data clustering & No & No & Categorical and numeric data support \\
\textbf{This Study (2025)} & K-means + Genetic Algorithm & Route planning for field maintenance & Yes & Yes (Istanbul) & Real-world, urban-scale application with operational focus \\ \bottomrule
\end{tabular}}
\end{table}

Although these studies offer strong methodological foundations, most are either simulation-based or focus on high-level infrastructure planning. Few have demonstrated the real-world applicability of these algorithms in the context of urban maintenance logistics. The current study addresses this gap by applying a well-established hybrid algorithm—K-means clustering followed by a genetic algorithm—for route optimization in a live dataset of 100 EV charging stations in Istanbul. Unlike previous work, this study emphasizes field-level operational planning within a densely populated, resource-constrained metropolitan environment, making it a practical contribution to EV infrastructure management.

\section{Methodology}
In this section, information about the dataset, methods and the proposed method used in the study is given.

\subsection{Materials}
This study employs a real-world dataset of electric vehicle (EV) charging stations located in Istanbul, Turkiye. The dataset includes geographic coordinates and identifier information for each station in Istanbul, forming the basis for clustering and route optimization tasks. 

\begin{figure}[htp]
    \centering
    \includegraphics[width=0.8\linewidth]{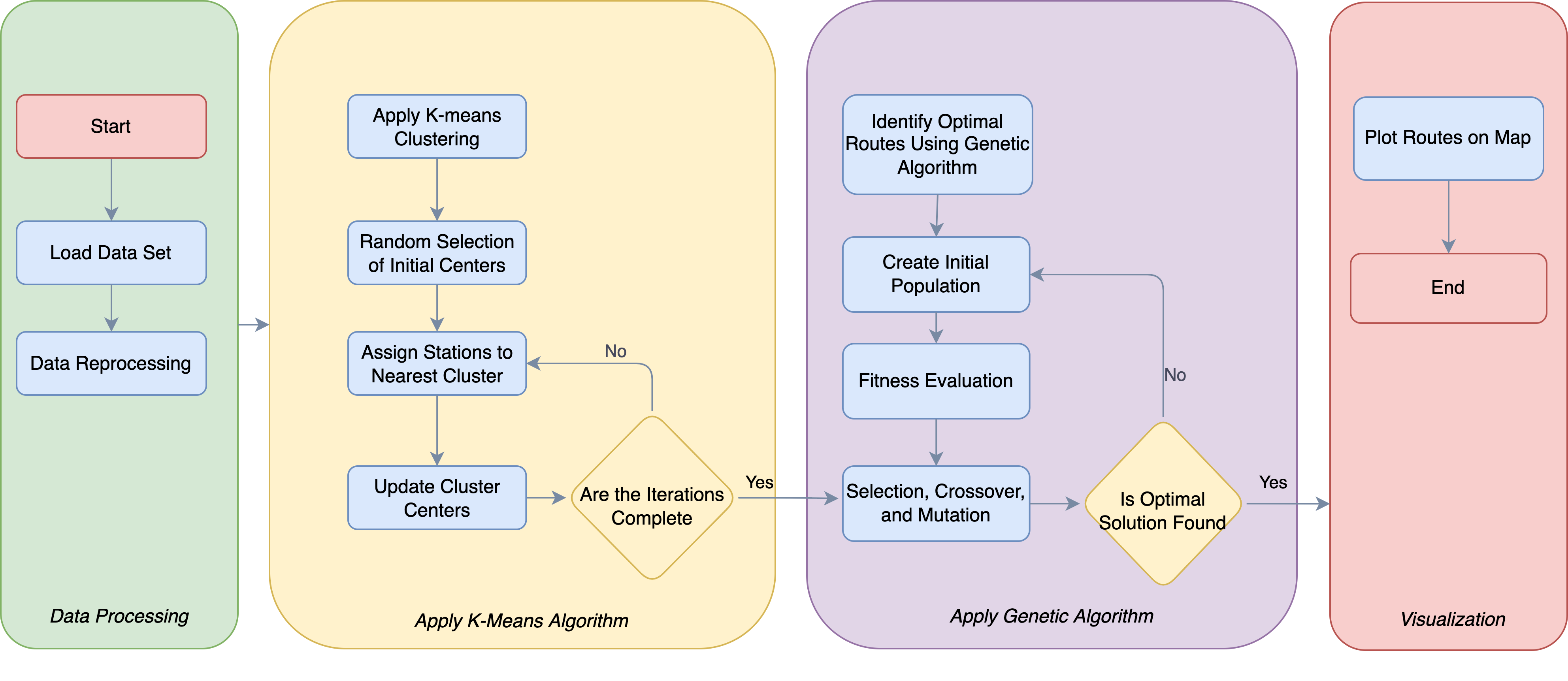}
    \caption{Proposed Algorithm}
    \label{fig:figure2}
\end{figure}

Figure~\ref{fig:figure2} shows the steps of the study that aims to determine the most efficient routes according to the locations of electric vehicle charging stations. 

\begin{itemize}
    \item \textbf{Dataset:} A collection of 100 EV charging stations, each with latitude, longitude, and station name.
    \item \textbf{Geospatial APIs:} Google Maps API and the \texttt{gmplot} Python library were used for mapping and visualization. Google Maps API also provided driving distance data between stations.
    \item \textbf{Programming Tools:} Python was used for data processing and algorithm implementation. Libraries such as Pandas, NumPy, math, random, and gmplot facilitated data manipulation, clustering, and plotting.
\end{itemize}

All codes and dataset related to the study have been shared via github repository \cite{yenigun2025route}.

\subsubsection*{Proposed Algorithms Pseudo Code}
Algorithm-1 shows the pseudo code of the proposed method.

\begin{algorithm}[htp]
\caption{EV Charging Point Optimization Pseudo code}
\begin{algorithmic}[1]

\STATE \textbf{Input:} Data = electric vehicle charging points
\STATE \textbf{Parameters:} max\_clusters, mutation\_rate, iterations

\STATE \textbf{Step 1: Data Preparation}
\STATE Load charging point coordinates into \texttt{data}

\STATE \textbf{Step 2: Clustering Charging Points with K-Means}
\STATE Initialize KMeans with \texttt{max\_clusters}
\STATE Fit KMeans model on \texttt{data}
\STATE Extract \texttt{cluster\_centers} and \texttt{clusters} from model

\STATE \textbf{Step 3: Finding Optimal Routes Within Each Cluster}
\STATE Initialize \texttt{routes} as empty list
\FOR{each \texttt{cluster\_id} in unique(\texttt{clusters})}
    \STATE \texttt{point\_list} $\leftarrow$ subset of \texttt{data} where \texttt{clusters == cluster\_id}
    \STATE \texttt{route} $\leftarrow$ determine optimal route within \texttt{point\_list} (e.g., using TSP)
    \STATE Append \texttt{route} to \texttt{routes}
\ENDFOR

\STATE \textbf{Step 4: Global Optimization for Inter-Cluster Routing}
\STATE Initialize Genetic Algorithm with \texttt{cluster\_centers}, \texttt{mutation\_rate}, \texttt{iterations}
\STATE \texttt{global\_route} $\leftarrow$ genetic algorithm.optimize()

\STATE \textbf{Step 5: Constructing the Final Route}
\STATE Initialize \texttt{final\_route} as empty list
\FOR{each \texttt{cluster\_index} in \texttt{global\_route}}
    \STATE \texttt{cluster\_route} $\leftarrow$ \texttt{routes[cluster\_index]}
    \STATE Append \texttt{cluster\_route} to \texttt{final\_route}
\ENDFOR

\STATE \textbf{Step 6: Visualization}
\STATE Plot \texttt{final\_route} and \texttt{cluster\_centers} on map
\end{algorithmic}
\end{algorithm}

The process starts with loading and reprocessing the dataset, and then using the K-Means algorithm, stations are assigned to the closest clusters according to randomly selected centers; this process is completed iteratively by updating the centers. After the clusters are determined, the optimal routes are found for each cluster by applying the genetic algorithm; at this stage, the initial population is created, the suitability is evaluated, and the cycle continues until the best solution is obtained by selection, crossover, and mutation processes. Finally, both intra-cluster and inter-cluster routes are combined and visualized on the map, and the process is completed.

\subsection{Methods}
To optimize field operation routes among geographically dispersed charging stations, a two-stage algorithm was implemented. K-means clustering was first used to group nearby stations based on spatial proximity. Genetic Algorithm (GA) then applied to optimize the order in which these station clusters should be visited.

\subsubsection{Data Preprocessing}
Initial processing involved structuring the dataset for clustering:

\begin{itemize}
    \item \textbf{Coordinates:} Latitude and longitude values were combined into coordinate pairs for spatial analysis.
    \item \textbf{Names:} Station names were stored in a separate list to retain reference information during clustering and mapping.
\end{itemize}

Table~\ref{tab:table2} presents 10 sample from the dataset.

\renewcommand{\arraystretch}{0.8} 
\begin{table}[htbp]
\centering
\caption{Example of the dataset}
\label{tab:table2}
\begin{tabular}{@{}lrr@{}}
\toprule
\textbf{Charging Point} & \textbf{Latitude} & \textbf{Longitude} \\ \midrule
Optimum Metro           & 40.91091997572458 & 29.296356708450354 \\
Nuvo Dragos             & 40.91753623170756 & 29.159728927618517 \\
GESAN                   & 40.94343325168699 & 29.137293687526036 \\
Up City Flats           & 40.93464386474212 & 29.215150617826865 \\
RES                     & 40.97945767531640 & 29.267345683206024 \\
Ucay Sultanbeyli Branch & 40.95864230497735 & 29.289594858306355 \\
Ritim Istanbul          & 40.96024022021866 & 29.159271365163416 \\
Marmara University      & 40.91091997572458 & 29.187367798227427 \\
Aydos                   & 40.96162180648609 & 29.219283032946160 \\
Bostanci                & 40.95655584884026 & 29.103615717225853 \\ \bottomrule
\end{tabular}
\end{table}

\subsubsection{K-Means Clustering}
\begin{figure}[H]
    \centering
    \scalebox{0.30}{\includegraphics{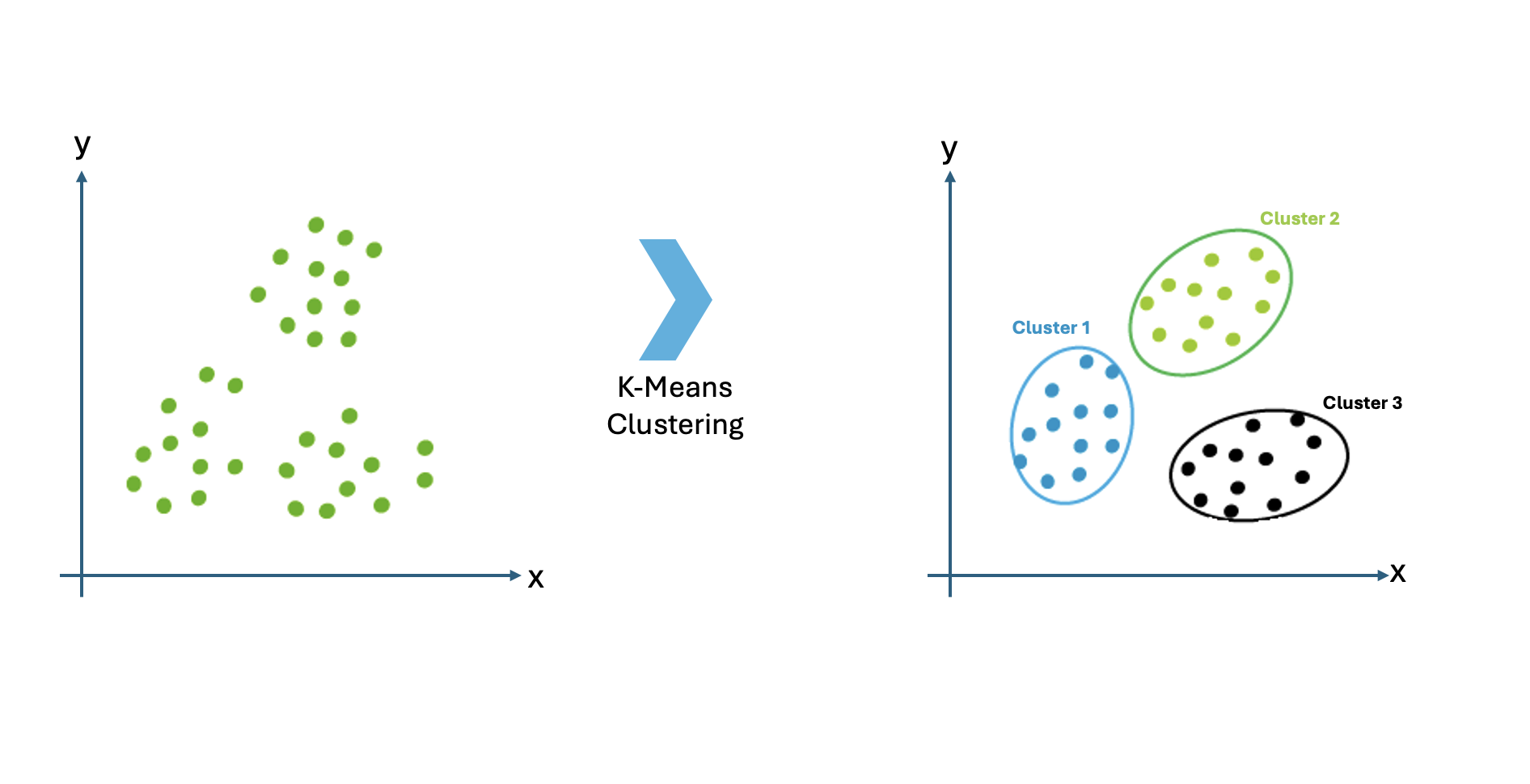}}
    \caption{K-Means Clustering Process}
    \label{fig:figure3}
\end{figure}

K-means clustering was used to divide stations into geographically coherent groups \cite{baydogmus2023}. This helps reduce intra-cluster travel and simplifies subsequent route optimization. Figure~\ref{fig:figure3} is an example representation of the K-means algorithm.

The algorithm proceeds as given steps \cite{Ikotun2023}:

\begin{enumerate}
    \item \textbf{Initialize Centroids:} Randomly select \textit{k} initial centroids.
    \item \textbf{Assign Points:} Assign each point to the nearest centroid using Euclidean distance as given in Equation~\ref{eq:equation1},
    \begin{equation}
        d(x,y)=\sqrt{(x_{1}-y_{1})^{2}+(x_{2}-y_{2})^{2}}
    \label{eq:equation1}
    \end{equation}
    \item \textbf{Update Centroids:} Compute the mean of all points in each cluster as given in Equation~\ref{eq:equation2},
    \begin{equation}
        C_{i}=\left( \frac{1}{n} \right) \sum x_{j}
    \label{eq:equation2}
    \end{equation}
    \item \textbf{Iterate:} Repeat the assignment and update steps until convergence or a fixed iteration limit is reached.
\end{enumerate}

As a result of these processes, stations located near each other are clustered and prepared for route optimization.

\subsubsection{Genetic Algorithm for Route Optimization}
After clustering, Genetic Algorithm (GA) was used to determine the optimal route between the identified station clusters \cite{mirjalili2019, kirtay2025genetic}. This heuristic, inspired by biological evolution, is suitable for solving complex combinatorial problems like the Traveling Salesman Problem (TSP) \cite{Cui2023, Ochelska-Mierzejewska2021, baydogmus2023}.

\begin{figure}[htp]
    \centering
    \scalebox{0.25}{\includegraphics{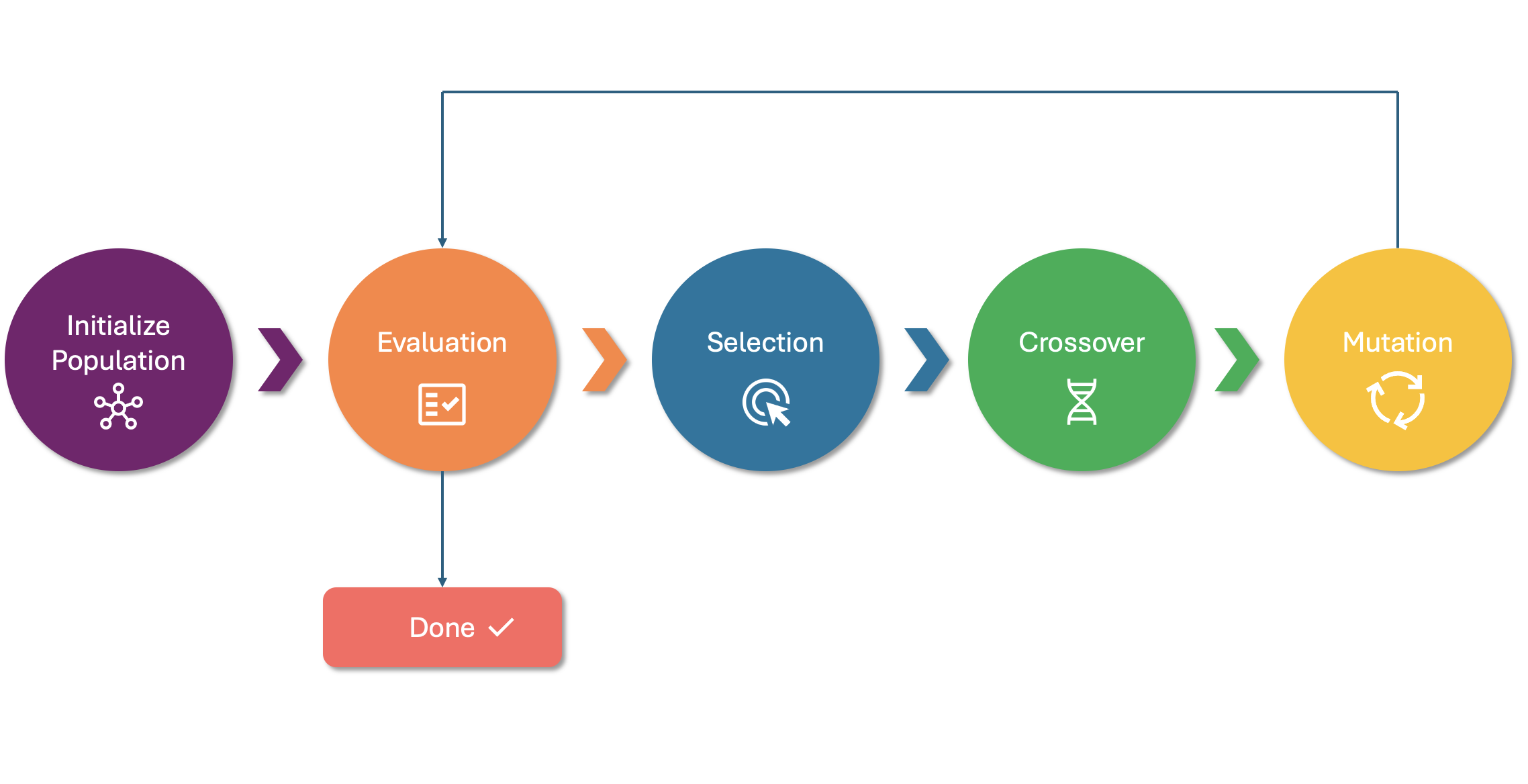}}
    \caption{Genetic Algorithm Workflow}
    \label{fig:figure4}
\end{figure}

GAs algorithm follows main steps given in  Figure~\ref{fig:figure4},

\begin{enumerate}
    \item \textbf{Population Initialization:} Random sequences of clusters were generated to form initial candidate routes.
    \item \textbf{Fitness Evaluation:} Total Euclidean distances were calculated for each route to determine fitness.
    \item \textbf{Selection:} Top-performing routes with shortest distances were selected for crossover.
    \item \textbf{Crossover:} New candidate routes were generated by combining parent routes.
    \item \textbf{Mutation:} Random alterations were introduced to preserve genetic diversity and avoid local optima.
\end{enumerate}

The genetic algorithm is terminated when a specified iteration limit is reached or the optimal solution is found \cite{karatacs2016genetic}. As a result, the shortest paths between clusters are determined, and optimal route planning is achieved \cite{Li2023, Alolaiwy2023}.

\subsubsection{Visualization and Mapping}

\begin{figure}[htp]
    \centering
    \scalebox{0.30}{\includegraphics{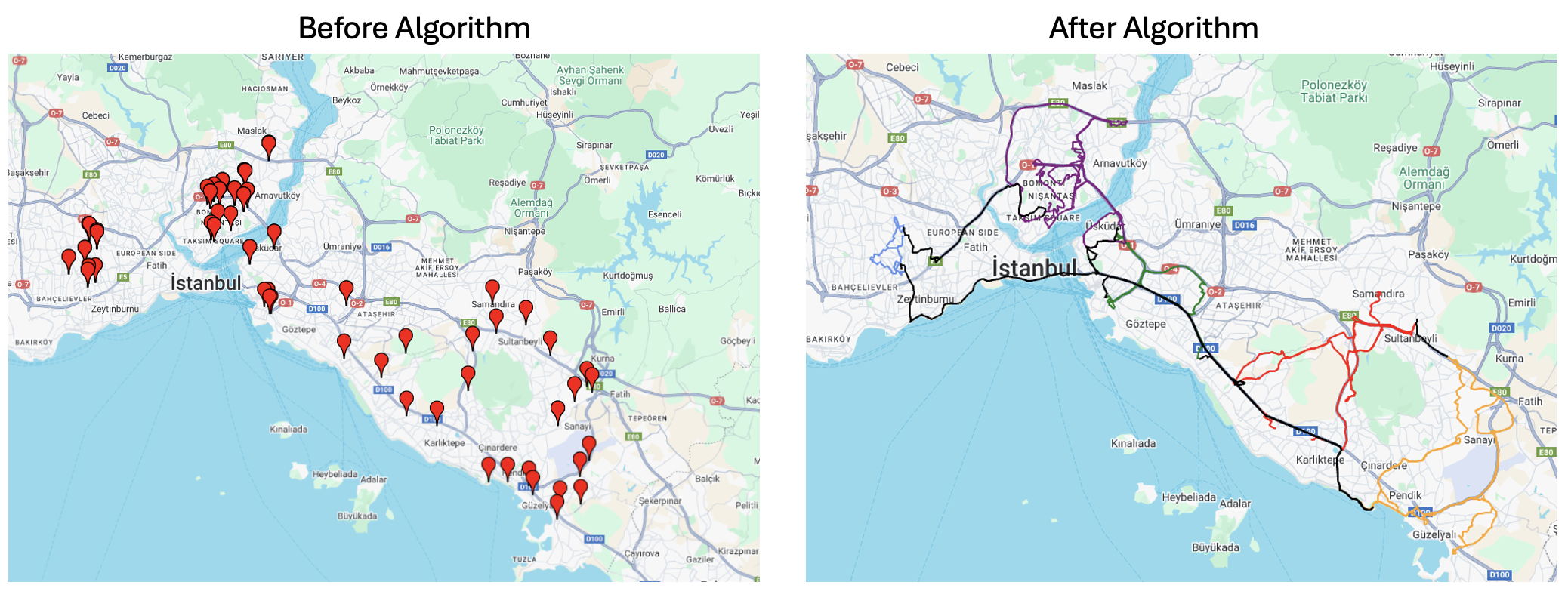}}
    \caption{Comparison: Pre-Optimization and Post-Optimization Routes}
    \label{fig:figure5}
\end{figure}

To illustrate the clustering and routing results,
\begin{itemize}
    \item K-means clusters were rendered in different colors.
    \item Genetic Algorithm results were visualized as connecting lines between clusters.
    \item Google Maps API was used for real road-based distance extraction.
    \item The final map was exported as an interactive HTML file for practical use by field operators.
\end{itemize}
Figure~\ref{fig:figure5} shows the results that will occur in the routes when K-means and GA are applied.

\subsubsection{Algorithm Parameters and Performance Evaluation}
To evaluate the performance of the algorithms used in this study, the number of K-means clusters (k), the number of iterations, and the Genetic Algorithm parameters were considered.

\begin{itemize}
    \item Number of K-means Clusters (k): The value of k was chosen variably based on station density and regional needs of field operations. In addition, the number of iterations was determined as a variable.
    \item Genetic Algorithm Parameters: Parameters such as number of iterations, and mutation rate were adjusted to optimize the route.
\end{itemize}

\section{Experimental Results}
This section presents a set of experiments conducted to evaluate the effectiveness of the proposed optimization framework. By systematically varying algorithmic parameters, the impact of clustering granularity and genetic algorithm behavior on route optimization was analyzed. Key parameters include the number of clusters, iteration count, mutation rate, and genetic algorithm iteration count. Table~\ref{tab:table2} summarizes the configurations tested and the corresponding result figures. More than 10 parameter combinations were made for the study, and the 4 most prominent ones were selected.

\renewcommand{\arraystretch}{1.0} 
\begin{table}[htbp]
\centering
\caption{Important parameters for proposed algorithm}
\label{tab:table3}
\begin{tabular}{@{}rrr r c@{}}
\toprule
\multicolumn{2}{c}{\textbf{K-means}} & \multicolumn{2}{c}{\textbf{GA}} & \multirow{2}{*}{\textbf{Result}} \\ \cmidrule(lr){1-2} \cmidrule(lr){3-4}
\textbf{Clusters} & \textbf{Iterations} & \textbf{Mutation Rate} & \textbf{Iterations} & \\ \midrule
3 & 10 & 0.10 & 100 & Fig. 5 \\
5 & 10 & 0.10 & 100 & Fig. 6 \\
5 & 10 & 0.05 & 200 & Fig. 7 \\
8 & 20 & 0.05 & 200 & Fig. 8 \\ \bottomrule
\end{tabular}
\end{table}

The use of a real dataset from Istanbul—a megacity characterized by high population density, irregular road layouts, and limited operational resources—adds significant practical value to this study. Unlike many simulation-based approaches in the literature, this work demonstrates the applicability of the proposed algorithm in an actual urban environment with complex spatial dynamics. 

\subsection{Configurations and Observations}
Various parameter and iteration experiments were conducted in the study. In this section, the most prominent parameters and the operations performed with them are explained.

\begin{figure}[htp]
        \centering
        \scalebox{0.45}{\includegraphics{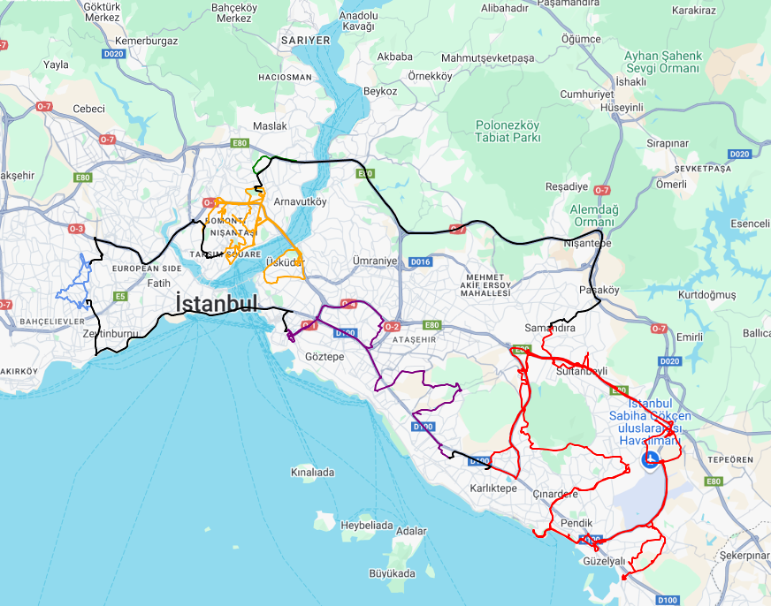}}
        	\caption{Configuration - 1 parameters and result}
            \label{fig:figure6}
        \end{figure}

\textbf{Configuration 1 – Broad Clustering, Moderate GA:}  
This configuration demonstrates the effect of fewer clusters on the routing process. By limiting the K-means clustering to 3 groups, the resulting routes focus on broader geographic areas, which simplifies the overall clustering but increases the travel distances within each cluster. The genetic algorithm, with a mutation rate of 0.1 and 100 iterations, provides a moderately optimized route between clusters. Result map is given in Figure~\ref{fig:figure6},

\begin{figure}[H]
        	\centering
        	\scalebox{0.45}{\includegraphics{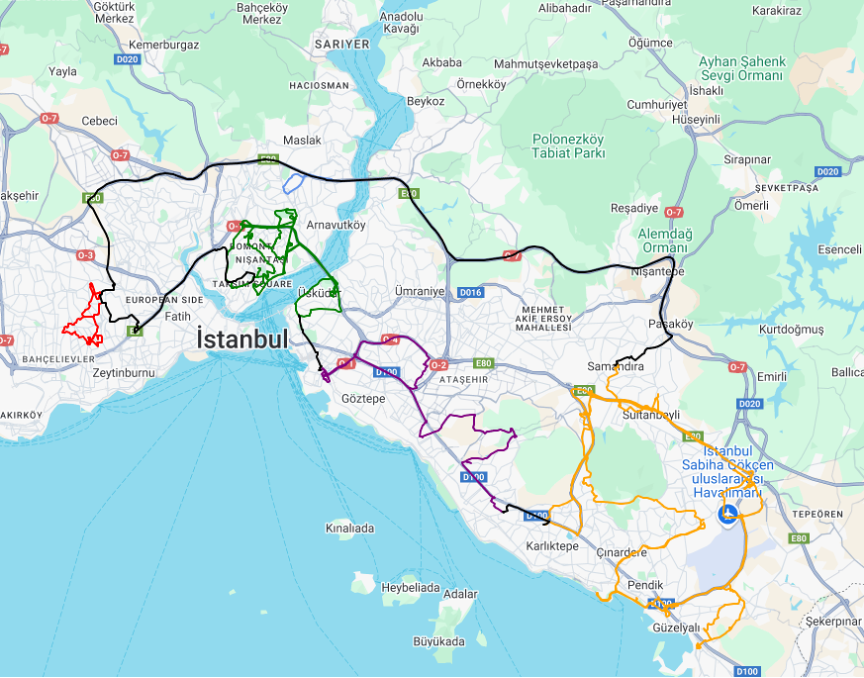}}
        	\caption{Configuration - 2 parameters and result}
            \label{fig:figure7}
        \end{figure}

\textbf{Configuration 2 – Finer Clustering, Same GA:}  
Increasing the number of clusters to 5 creates a more granular segmentation of charging stations. This reduces the size of each cluster, resulting in shorter intra-cluster travel distances. However, the mutation rate and iteration count remain the same, leading to a route optimization level similar to the previous configuration but with improved efficiency due to the refined clustering. Result map is given in Figure~\ref{fig:figure7},

\begin{figure}[H]
        	\centering
        	\scalebox{0.45}{\includegraphics{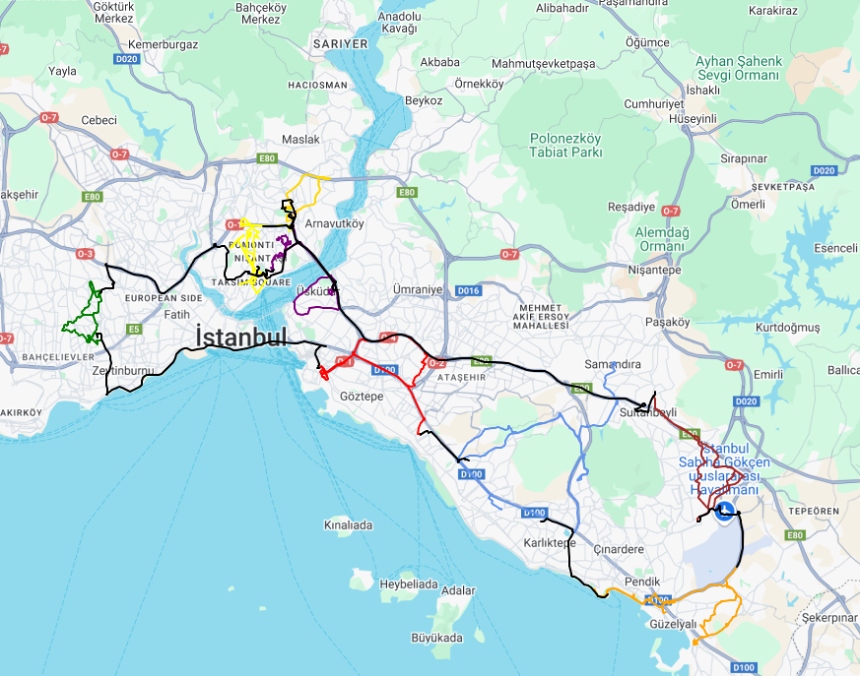}}
        	\caption{Configuration - 3 parameters and result}
            \label{fig:figure8}
        \end{figure}

\textbf{Configuration 3 – Finer Clustering, Enhanced GA:}  
Iteration: 10, Genetic Algorithm Mutation Rate: 0.05, Genetic Algorithm Iteration: 200
    Maintaining 5 clusters but lowering the genetic algorithm’s mutation rate to 0.05 and increasing iterations to 200 enhances the route optimization process. This configuration achieves more precise routes by allowing the algorithm to converge towards better solutions, balancing exploration and exploitation, and yielding reduced total travel distances. Result map is given in Figure~\ref{fig:figure8},

\textbf{Configuration 4 – High Granularity and Stability:}  
This configuration maximizes clustering granularity by dividing the charging stations into 8 groups, providing highly localized clusters. The increased K-means iterations improve cluster stability, while the genetic algorithm, with lower mutation and higher iteration parameters, delivers highly optimized routes. This setup results in minimal intra-cluster and inter-cluster travel distances, representing the most refined and efficient routing solution.

The complexity and variability inherent in Istanbul's urban geography make efficient routing especially challenging. The consistent performance of the proposed optimization method across multiple configurations illustrates its adaptability to dense metropolitan contexts. Such robustness is particularly valuable for charge point operators in cities where service delays and operational inefficiencies are magnified by traffic, scale, and spatial distribution.

\subsection{Comparison Between Reference Route and Optimized Route}
To assess performance, the optimized result from Configuration - 4 was compared to a baseline “reference route” generated by traversing charging stations sequentially as they appear in the dataset. No special optimization was applied to the reference route.

Distance calculation and evaluation metrics for distance computation and improvement evaluation is outlined,
\begin{itemize}
    \item \textbf{Reference Route Distance:} Summation of Euclidean distances between consecutive data points in dataset order.
    \item \textbf{Optimized Route Distance:} Sum of intra-cluster distances from K-means and inter-cluster distances from the Genetic Algorithm.
    \item \textbf{Improvement Rate:}
    \begin{equation}
        \text{Improvement (\%)} = \frac{\text{Reference Distance} - \text{Optimized Distance}}{\text{Reference Distance}} \times 100
    \end{equation}
\end{itemize}

The evaluations were made by considering the given calculations, reference route and optimized route distances obtained with the hybrid approach. Since used dataset belongs to the real world, the reference distance was also needed. 

In line with this configurations, Table~\ref{fig:figure4} shows the distance and running time information obtained as a result of certain optimizations.
\renewcommand{\arraystretch}{1.0} 
\begin{table}[htbp]
\centering
\caption{Route distance and duration}
\label{tab:table4}
\begin{tabular}{@{}rrrr@{}}
\toprule
\textbf{Config No} & \textbf{K-Means Cluster Count} & \textbf{Total Distance (km)} & \textbf{Total Duration (min)} \\ \midrule
1 & 3 & 330.55 & 615.37 \\
2 & 5 & 308.49 & 568.15 \\
3 & 5 & 314.46 & 586.65 \\
4 & 8 & 359.75 & 634.18 \\
Sequential Route & 100 (points) & 417.53 & 706.22 \\ \bottomrule
\end{tabular}
\end{table}

\begin{figure}[H]
        	\centering
        	\scalebox{0.20}{\includegraphics{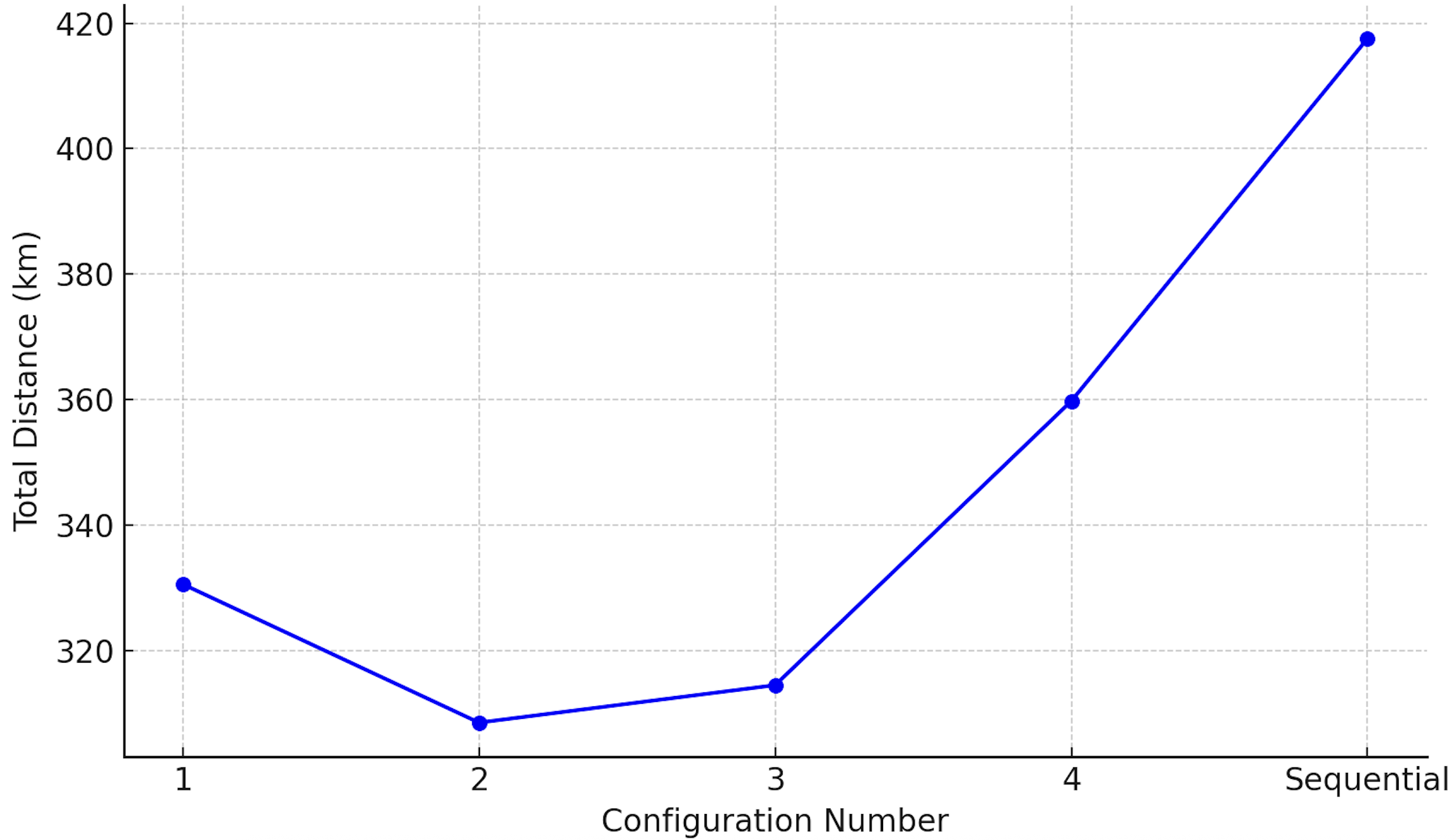}}
        	\caption{Total distance comparison}
            \label{fig:figure9}
\end{figure}

Figure~\ref{fig:figure9} shows the total distance obtained by clusters more clearly. Accordingly, the shortest distance was obtained with configuration-2 (308.49 km). The longest distance was reached in the "Sequential" route (417.53 km), which shows that a non-optimized layout is less efficient. In other words, optimized (using K-Means) configurations cover less distance compared to the sequential route.

\begin{figure}[H]
        	\centering
        	\scalebox{0.20}{\includegraphics{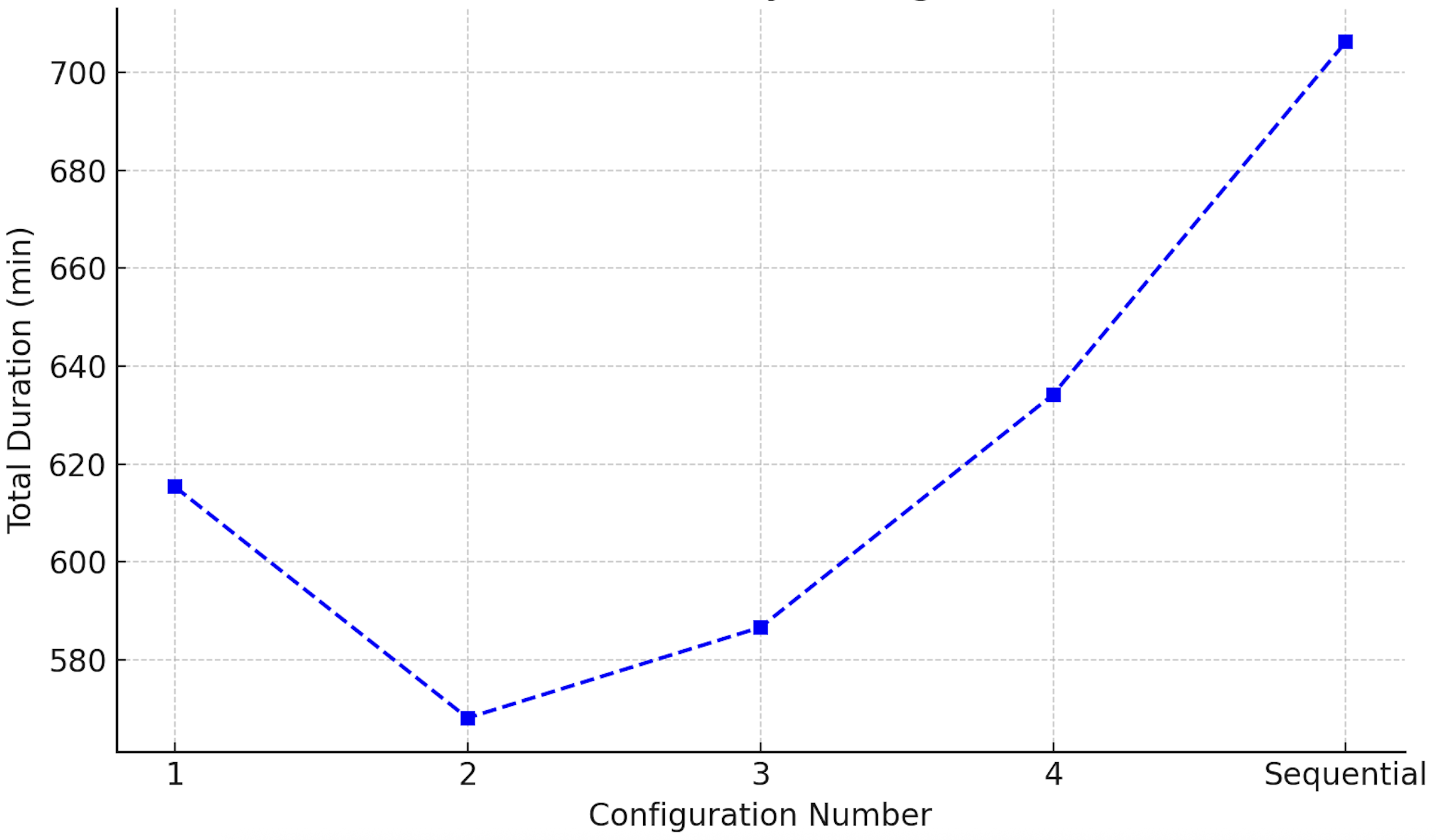}}
        	\caption{Total duration comparison}
            \label{fig:figure10}
\end{figure}

In addition, Figure~\ref{fig:figure10} shows the total distance obtained by different clusters. Accordingly, the shortest time was obtained in configuration-2 (568.15 minutes). This configuration also allowed us to obtain the most efficient distance. The "Sequential" route is again completed in the longest time (706.22 minutes), which shows that inefficiency is reflected in both time and distance. In other words, configurations optimized with K-means seem to be quite advantageous in terms of time.

The comparison showed that the optimized route reduced total travel distance by approximately 35\%. Importantly, this is not only a numerical improvement but also a reflection of the method’s ability to handle real-world complexity in a city like Istanbul. In such urban environments, every kilometer saved translates into tangible operational savings, reduced technician workload, and improved service continuity.

These findings highlight the significant reduction in distance achieved by optimization algorithms compared to a non-optimized reference route. They also underscore the practical benefits of optimization techniques. However, since the reference route is fully dependent on the dataset's ordering, the improvement percentage may vary with different orderings and should be regarded as an approximate value.

\section{Discussion}
The results of this study demonstrate that combining K-means clustering and GA provides a practical and effective solution for route optimization in EV charging station maintenance. The observed reductions in total travel distance across varying parameter configurations affirm the adaptability of the proposed approach to different operational conditions.

A key strength of this work lies in its use of real spatial data from Istanbul—a city known for its complex topology, heavy traffic, and high population density. This distinguishes the study from prior research that primarily relies on synthetic or abstract datasets. The ability of the optimization framework to operate efficiently under such conditions confirms its value for deployment in real-world metropolitan contexts, particularly for charge point operators (CPOs) aiming to improve field resource management.

The comparative literature analysis presented in Table~\ref{tab:table1} further reinforces the novelty of this work. While previous studies have made valuable contributions in EV infrastructure planning, most are simulation-based or focus on high-level strategic placement of stations. Few have addressed the operational dimension of maintenance routing under real-world constraints. Moreover, as the table highlights, most reviewed methods lack application in real urban environments. In contrast, this study offers a field-deployable solution grounded in real data from Istanbul—a geographically complex and logistically demanding city. This contextual uniqueness adds not only practical relevance but also academic value by bridging the gap between theoretical optimization models and their real-world applicability.

Despite the positive outcomes, certain limitations must be acknowledged. The use of Euclidean distance, while computationally convenient, does not account for real road networks or traffic data. Additionally, the baseline “reference route” is a naive comparator based on dataset order, which may not reflect realistic manual planning. As such, the reported 35\% improvement should be viewed as indicative rather than absolute. Further benchmarking against heuristic baselines like greedy nearest-neighbor or 2-opt algorithms would strengthen comparative analysis.

Moreover, this study models a static routing scenario. In real-world applications, dynamic factors such as technician availability, traffic conditions, time windows, and service priority levels may influence routing decisions. Integrating these constraints into future models would enhance realism and applicability.

Overall, the study contributes to the practical literature by validating a lightweight, data-driven route planning method that is readily deployable in urban maintenance contexts. Its success in a high-density environment like Istanbul suggests potential scalability to other large cities with similar logistical constraints.

\section{Conclusion and Future Work}
This study demonstrates that integrating K-means clustering with a Genetic Algorithm offers an effective and scalable solution for optimizing field operations in electric vehicle (EV) charging networks. By combining spatial clustering with evolutionary route optimization, the proposed framework reduces operational travel distance, thereby lowering costs, minimizing resource consumption, and improving overall efficiency.

A key contribution of this work lies in its application to real geospatial data from Istanbul—a densely populated and logistically complex urban environment. Unlike previous studies based on synthetic data or simulation models, this research validates its approach in a real-world metropolitan setting. This adds both practical and academic value, demonstrating how established algorithms can be operationalized to solve modern urban mobility challenges.

The method’s performance across multiple parameter configurations also underscores its adaptability to different spatial and operational conditions. The findings suggest that hybrid clustering-routing algorithms are well-suited for charge point operators (CPOs) seeking to streamline maintenance activities and ensure service continuity.

Future work may extend this framework by incorporating real-time traffic data, technician schedules, service priority levels, and dynamic routing constraints. Additionally, benchmarking against other heuristic baselines and testing in other large cities could further validate the model’s scalability and generalizability. Investigating the integration of more advanced metaheuristics or reinforcement learning techniques may also unlock further performance gains in large-scale deployments.

\bibliographystyle{unsrt}  
\bibliography{references}

\begin{thebibliography}{10}

\bibitem{Vasiliki2023}
Vasiliki Lazari.
\newblock Multi-objective optimization of electric vehicle charging station deployment using genetic algorithms.
\newblock {\em MDPI}, 13(8), 2023.

\bibitem{Akbari2018}
Milad Akbari.
\newblock Optimal locating of electric vehicle charging stations by application of genetic algorithm.
\newblock {\em Sustainability}, 10, 2018.

\bibitem{iea_ev_2025}
{International Energy Agency}.
\newblock Electric vehicles -- iea.
\newblock \url{https://www.iea.org/energy-system/transport/electric-vehicles}, 2025.
\newblock Accessed: January 10, 2025.

\bibitem{Masoud2018}
Mahmoud Masoud.
\newblock A simulated annealing for optimizing assignment of e-scooters to freelance chargers.
\newblock {\em Sustainability}, 10, 2018.

\bibitem{Kumar2024}
Boya~Anil Kumar.
\newblock Hybrid genetic algorithm-simulated annealing based electric vehicle charging station placement for optimizing distribution network resilience.
\newblock {\em Scientific Reports}, 14, 2024.

\bibitem{optimize_non_linear_ev_routing}
S.~Karakatic.
\newblock Optimizing nonlinear charging times of electric vehicle routing with genetic algorithm.
\newblock {\em Expert Systems with Applications}, 164, 2021.

\bibitem{ev_charging_infra_location}
Dimitrios Efthymiou and K.~C. M. M.~G. A.
\newblock Electric vehicles charging infrastructure location: a genetic algorithm approach.
\newblock {\em Springer}, 2017.

\bibitem{location_opt_ev_stations}
Guangyou Zhou and Z.~Z.~S. L.
\newblock Location optimization of electric vehicle charging stations: Based on cost model and genetic algorithm.
\newblock {\em Energy}, 247, 2022.

\bibitem{opt_kmeans_ga}
Shadab Irfan and G.~D.~S. G.
\newblock Optimization of k-means clustering using genetic algorithm.
\newblock In {\em International Conference on Computing and Communication Technologies for Smart Nation}, 2017.

\bibitem{kmeans_ga_review}
Diyar~Qader Zeebaree and H.~H. A. M. A. S. R.~M. Z.
\newblock Combination of k-means clustering with genetic algorithm: A review.
\newblock {\em International Journal of Applied Engineering Research}, 12(24), 2017.

\bibitem{optimized_kmeans_bat}
A.~W.~G. Komarasamy.
\newblock An optimized k-means clustering technique using bat algorithm.
\newblock {\em European Journal of Scientific Research}, 84:263--273, 2012.

\bibitem{genetic_kmeans_mixed}
Dharmendra~K. Roy and L.~K. S.
\newblock Genetic k-means clustering algorithm for mixed numeric and categorical data sets.
\newblock {\em International Journal of Artificial Intelligence and Applications}, 1(2), 2010.

\bibitem{yenigun2025route}
Onur Yenigun.
\newblock Route optimization.
\newblock \url{https://github.com/oyenigun/RouteOptimization}.
\newblock Accessed: 2025.

\bibitem{baydogmus2023}
G.~K. Baydogmus.
\newblock Solution for tsp/mtsp with an improved parallel clustering and elitist aco.
\newblock {\em Computer Science and Information Systems}, 20(1):195--214, 2023.

\bibitem{Ikotun2023}
Abiodun~M. Ikotun.
\newblock K-means clustering algorithms: A comprehensive review, variants analysis, and advances in the era of big data.
\newblock {\em Elsevier}, 622:178--210, 2023.

\bibitem{mirjalili2019}
S.~Mirjalili and S.~Mirjalili.
\newblock Genetic algorithm.
\newblock In {\em Evolutionary Algorithms and Neural Networks: Theory and Applications}, pages 43--55. 2019.

\bibitem{kirtay2025genetic}
Seda Kirtay, Veysel~Gokhan Bocekci, and Kazim Yildiz.
\newblock Genetic algorithm based approach for optimization of fair power allocation in pd-noma systems.
\newblock {\em Tehni{\v{c}}ki vjesnik}, 32(3):948--957, 2025.

\bibitem{Cui2023}
Huixia Cui.
\newblock Route optimization in township logistics distribution considering customer satisfaction based on adaptive genetic algorithm.
\newblock {\em Elsevier}, 204:28--42, 2023.

\bibitem{Ochelska-Mierzejewska2021}
Joanna Ochelska-Mierzejewska.
\newblock Selected genetic algorithms for vehicle routing problem solving.
\newblock {\em MDPI}, 10, 2021.

\bibitem{karatacs2016genetic}
G{\"o}zde Karata{\c{s}}.
\newblock Genetic algorithm for intrusion detection system.
\newblock In {\em 2016 24th Signal Processing and Communication Application Conference (SIU)}, pages 1341--1344. IEEE, 2016.

\bibitem{Li2023}
Chunhui Li.
\newblock Route optimization of electric vehicles based on reinsertion genetic algorithm.
\newblock {\em IEEE Transactions on Transportation Electrification}, 9(3), 2023.

\bibitem{Alolaiwy2023}
Muhammad Alolaiwy.
\newblock Multi-objective routing optimization in electric and flying vehicles: A genetic algorithm perspective.
\newblock {\em MDPI}, 13(18), 2023.

\end{thebibliography}

\end{document}